# Applying Fine-Tuned LLMs for Reducing Data Needs in Load Profile Analysis

Yi Hu, Hyeonjin Kim, Kai Ye, Ning Lu

*Abstract*—This paper presents a novel method for utilizing fine-tuned Large Language Models (LLMs) to minimize data requirements in load profile analysis, demonstrated through the restoration of missing data in power system load profiles. A two-stage fine-tuning strategy is proposed to adapt a pre-trained LLMs, i.e., GPT-3.5, for missing data restoration tasks. Through empirical evaluation, we demonstrate the effectiveness of the fine-tuned model in accurately restoring missing data, achieving comparable performance to state-of-the-art specifically designed models such as BERT-PIN. Key findings include the importance of prompt engineering and the optimal utilization of fine-tuning samples, highlighting the efficiency of few-shot learning in transferring knowledge from general user cases to specific target users. Furthermore, the proposed approach demonstrates notable cost-effectiveness and time efficiency compared to training models from scratch, making it a practical solution for scenarios with limited data availability and computing resources. This research has significant potential for application to other power system load profile analysis tasks. Consequently, it advances the use of LLMs in power system analytics, offering promising implications for enhancing the resilience and efficiency of power distribution systems.

## I. INTRODUCTION

MACHINE learning has been extensively used in power system load profile analysis tasks, including missing data restoration, load forecasting, load disaggregation, and load profile generation. Training a large machine learning model from scratch typically requires substantial data and considerable training time. However, due to security and privacy concerns, utilities cannot share vast amounts of load data with the research community for these analyses. As an alternative, synthetic load profiles derived from real datasets have been used [1]. However, this approach also requires significant effort in training generative machine learning models.

In this paper, we propose using fine-tuned Large Language Models (LLMs) for load profile analysis to address the challenges of data scarcity while reducing computational time and cost. Initially trained on vast, diverse datasets, LLMs develop an extensive understanding of patterns and relationships within data. By fine-tuning these models on specific tasks like load profile analysis, we leverage their pre-existing knowledge, reducing the need for large, task-specific datasets.

Fine-tuning is data-efficient, requiring only a small dataset to adapt the model to specific tasks—a significant advantage when data is scarce, sensitive, or expensive. This approach also demands fewer computational resources than training a model from scratch since it only adjusts a portion of the model's weights and is performed on smaller datasets [2].

Furthermore, fine-tuned LLMs offer the flexibility to adapt to various downstream tasks without extensive reprogramming, allowing the same base model to be used for different analysis tasks within the same field, such as load forecasting or missing data restoration. This flexibility ensures high levels of performance even when large, comprehensive datasets are unavailable due to privacy, security, or practicality concerns.

In this paper, we will demonstrate how to fine-tune LLMs for conducting the missing data restoration task, illustrating both the process and the benefits of our approach.

### A. Missing Data Restoration

Missing data is a common occurrence in load profile processing within power systems. These data gaps often stem from temporary equipment malfunctions or deliberate human interventions. Equipment malfunctions lead to periods without consumption measurements, resulting in missing data segments. Conversely, operations like demand response (DR) and conservation voltage reduction (CVR) do not directly interrupt consumption measurements but make the baseline consumption level unknown. This baseline, representing the consumption level if the operation were not implemented, constitutes another form of missing data. The absence of load data significantly impacts power distribution system analysis, affecting tasks such as modeling load behaviors, studying renewable integration, and designing demand response programs. Therefore, developing an algorithm to restore missing data has become a crucial solution to these challenges.

In the field of load profile inpainting, methods for restoring missing data can be broadly classified into two categories: *model-based* and *data-driven* approaches. Model-based methods utilize physical system models to simulate the missing data segments. For example, to estimate the CVR baseline,

This material is based upon work supported by the U.S. Department of Energy's Office of Energy Efficiency and Renewable Energy (EERE) under the Solar Energy Technologies Office. Award Number: DE-EE0008770.

Yi Hu, Hyeonjin Kim, Kai Ye, and Ning Lu are with the Electrical & Computer Engineering Department, Future Renewable Energy Delivery and Management (FREEDM) Systems Center, North Carolina State University, Raleigh, NC 27606 USA. (emails: yhu28@ncsu.edu, hkim66@ncsu.edu, kye3@ncsu.edu, nlu2@ncsu.edu).

TABLE I
APPLICATIONS OF LARGE LANGUAGE MODELS IN DIFFERENT AREAS OF RESEARCH

| Field of application | | Tasks | References | Main contribution |
|---|---|---|---|---|
| Engineering | Software engineering | Code generation | [29] | Investigating example selection for In-Context Learning to generate code effectively. |
| | | | [30] | Propose a novel prompting technique to improve the performance of LLMs in code generation. |
| | | Self-Debug | [31] | Identify the mistakes in code by investigating the execution results and explaining the generated code in natural language and streamline the debugging process. |
| | Mechanical engineering | Calculations in mechanical engineering | [32] | Encountered incorrect procedures, formulas, or results. Showing that ChatGPT should not be relied upon solving practical mechanical problems. |
| | Mathematical engineering | Mathematical calculations | [33] | Investigate the mathematical capabilities of ChatGPT by testing it on publicly available datasets. |
| | | Mathematical education | [34] | Use ChatGPT as a tool for teaching and learning mathematics |
| | Manufacturing | Support design, manufacturing, and engineering education | [35] | Propose a technology development roadmap to successfully integrate ChatGPT into the manufacturing industry. |
| | | Manufacturing troubleshooting. | [36] | Evaluated ChatGPT's ability in technical matters, including printing parameters and bed detachment, warping, and stringing issues. |
| Finance | Risk assessment | Risk assessment, algorithmic trading, and low-code development. | [37] | Propose an end-to-end framework, FinGPT, to serve as a catalyst for stimulating innovation in the finance domain. |
| Medical | medical education | Assistant learning and problem solving. | [38] | Shows ChatGPT's performance in the United States Medical Licensing Exam (USMLE) was comparable to or exceeded the passing threshold, indicating its proficiency in medical knowledge without requiring specialized training or reinforcement. |
| | radiologic decision-making | AI-based clinical decision-making | [39] | Highlighting its feasibility and potential benefits in improving clinical workflow and ensuring responsible use of radiology services |
| | clinical genetics | Answer genetics-related questions | [40] | ChatGPT's performance did not significantly differ from humans when answering genetics-related questions. However, the model demonstrated better accuracy on memorization-type questions compared to questions requiring critical thinking. |
| | Patient care | Gathering patient data, administering surveys or questionnaires. | [41] | The recent popularity of LLMs can potentially not only improve patient confidence in interacting with such chatbots but also improve upon the services provided. |

researchers use distribution system topology and load models to predict load variations in response to voltage changes [3]-[7]. These methods depend on accurate physics-based models that require numerous inputs and careful parameter tuning.

The data-driven approach provides a comprehensive solution for restoring missing data and has gained significant popularity due to its convenience. Researchers using this approach often categorize load profiles based on factors such as day type, weather conditions, and shape characteristics, then restore missing data segments by referencing load profiles with the closest similarity [8]-[11]. These methods are straightforward, easy to implement, and explainable. However, they often depend on subjective selections of similarity metrics and weights, typically defined by human analysts using weighted averages of various factors, making the accuracy contingent on the analyst's choices.

Recently, regression models have emerged as viable alternatives for load profile inpainting. Models such as Linear Regression [12], Long Short-term Memory networks (LSTM) [13], Stacked Autoencoder (SAE) [14], Gaussian Regression [15], Support Vector Regression (SVR) [16][17] have been increasingly employed to address the missing data restoration problem. These models generally achieve higher estimation accuracy compared to similarity-based methods due to their ability to capture nonlinear patterns in the data. However, it is important to note that deep-learning-based methods are less explainable and incur higher computational costs compared to similarity-based approaches. In recent years, hybrid solutions that combine multiple regression models have been proposed for baseline estimation and missing data restoration [18]-[20].

Additionally, GAN-based methods have also been utilized for restoring missing load data [21]-[27] and outperform model-based, similarity-based, regression-based, and other GAN-based methods in restoring missing data segments (MDSs) of fixed length. Moreover, a Bidirectional Encoder Representations from Transformers (BERT) based method is also employed to restore missing data in a daily load profile [28]. This method outperforms all the existing load inpainting methods in accuracy and can produce multiple data restoration candidates with varying confidence levels. Consequently, we select it as the benchmark method in this paper.

*B. Application of LLM in Different Areas of Research*

However, all the aforementioned methods are trained from scratch, necessitating a large amount of training data and extended training time. To address this, we propose solving the load profile inpainting problem by fine-tuning a Large Language Model (LLM). Fine-tuning is a powerful technique that leverages the pre-trained knowledge and general language understanding already present in the LLM. This allows for efficient adaptation to new tasks or domains with relatively small amounts of additional data. Fine-tuning has been widely used to customize language models for various applications. Table I provides a list of the main applications of LLMs in different areas of research.

Although LLMs have been applied to various fields, their application in power systems is still relatively limited. LLMs have been utilized in smart grid cybersecurity studies [42][43], and J. Ruan et al. have discussed the potential security threats associated with applying LLMs to power system [44]. However, research on using LLMs to restore missing data in power systems and estimate DR baselines is lacking. Investigating how LLMs perform compared to specifically designed missing data restoration models is a promising area of research.

In this paper, we propose a two-stage strategy to fine-tune the ChatGPT to minimize data requirements in load profile analysis, demonstrated through the restoration of missing data in power system load profiles. The contributions are three-fold.

- **Pioneering Application:** To the best of our knowledge, this is the first application of a fine-tuned LLM to a specific power system data analytics task, specifically missing data restoration and baseline estimation. The fine-tuned model achieves acceptable accuracy in missing data restoration, comparable to state-of-the-art specifically designed models.
- **Innovative Fine-Tuning Strategy:** We introduce a two-stage fine-tuning strategy using few-shot transfer learning. By leveraging a very small amount of training data, we can transfer the knowledge from a general language model to the task of power system missing data restoration. This approach significantly reduces the data needs and training time for load profile analysis.
- **Empirical Evaluation:** We evaluate the fine-tuned model using real smart meter data collected from various households and conduct a comprehensive study on the effect of different fine-tuning strategies. This offers valuable insights, including the importance of prompt engineering, optimal utilization of fine-tuning samples, and the efficiency of few-shot learning. These findings enhance understanding of model behavior and inform other load profile analysis tasks.

The rest of the paper is organized as follows. Section II introduces the fine-tuning methodology, Section III introduces the simulation results in different cases, and Section IV concludes the paper.

## II. METHODOLOGY

This section begins by formulating the missing load profile restoration problem. We then provide an overview of the proposed two-stage fine-tuning strategy, detailing its key components and functionalities. Additionally, we introduce performance evaluation metrics that will be used as benchmarks to assess the accuracy of the fine-tuned model.

### A. Missing Data Restoration Problem Formulation

In this context an *event* refers to a missing data segment (MDS), visually depicted as the green segment in Figure 1. The event is characterized by the starting time $T_{start}$ and the ending time $T_{end}$. Let $\boldsymbol{X} = [x_1, x_2, ..., x_N]$ represent a historical time series of load, where $N$ denotes the length of the time series. Consequently, $\boldsymbol{X}$ can be divided into three distinct parts:

$$\boldsymbol{X} = [\boldsymbol{X}_{pre}, \boldsymbol{X}_{event}, \boldsymbol{X}_{post}] \quad (1)$$

where $\boldsymbol{X}_{event} = [x_{T_{start}}, ..., x_{T_{end}}]$.

The objective of missing data restoration problem is to recover the missing portion of the load, denoted as $\boldsymbol{X}_{event}$, utilizing the pre-event load $\boldsymbol{X}_{pre}$ and post-event load $\boldsymbol{X}_{post}$. In essence, the missing data imputation problem can be defined as follows:

$$\widehat{\boldsymbol{X}}_{event} = f(\boldsymbol{X}_{pre}, \boldsymbol{X}_{post}) \quad (2)$$

where $f$ is the mapping function.

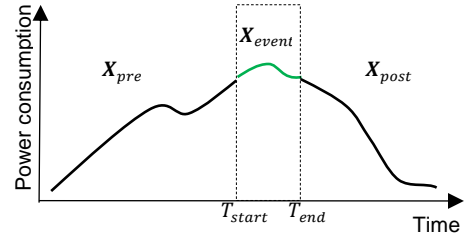

Fig. 1. Illustration of missing data imputation problem.

Moreover, the *event* can also be an unknown Demand Response (DR) baseline or Conservation Voltage Reduction (CVR) baseline. Because the DR or CVR baseline estimation is to recover the would-have-been load profiles assuming the DR or CVR program had not been executed, which can be considered as a special case of the missing data imputation problem. Thus, the missing data imputation method proposed in this paper can be applied to DR or CVR baseline estimation, which is valuable for load service providers to evaluate the performance of DR or CVR.

### B. Fine-tuning Strategy Overview

Although many machine learning based methods can be used to restore missing data in load profiles, we propose this Large Language Model fine-tuning method for two key reasons: First, our method requires significantly less data and training time compared to models trained from scratch, yet it achieves comparable accuracy in missing data restoration. Second, this approach eliminates the need to design complex models and training processes. Instead, we only need to develop a fine-tuning strategy based on the well-known ChatGPT-3.5 model.

The overview of the two-stage fine-tuning strategy is illustrated in Figure 2. In the first stage, the standard GPT-3.5 model is fine-tuned using data collected from 10 users similar to the target user. An encoding method and prompt technique are employed to create the intermediate model, GPT-FT-1. In the second stage, we further fine-tune this intermediate model using data from the target user. Different sample sizes are tested in this stage to achieve the best fine-tuning results.

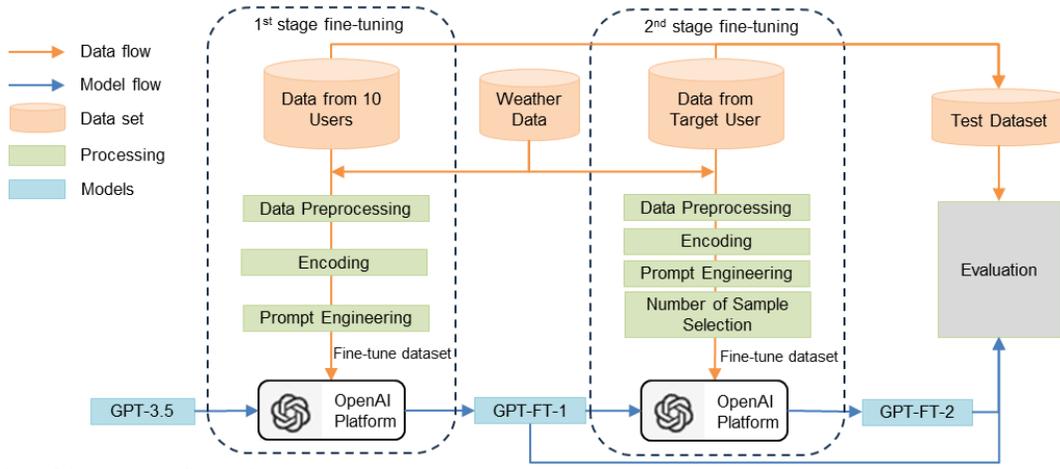

Fig. 2. An overview of the two-stage fine-tuning strategy.

## C. 1st Stage Fine-tuning

In this section, we will outline the main steps involved in the first stage of fine-tuning. These steps include data preprocessing, smart meter data-to-words encoding, prompt engineering, and OpenAI fine-tuning.

### 1) Data Preprocessing

Unlike the BERT-PIN model [28] which utilizes data from 8000 users over a three-year period, our first stage fine-tuning uses data collected from only 10 users over three months. The load profiles are segmented into daily load profiles for this process. To generate the time series data with MDS, we create a mask vector $M$ for each $X$, so that

$$M = [m_i \ for \ i = 1: N], m_i = \begin{cases} 0, & missing \ data \\ 1, & otherwise \end{cases} \quad (3)$$

Then, the masked load profile, $X^m$, can be represented as

$$X^m = X \cdot M \quad (4)$$

Note that we set all missing data segments to be 0 kW because all the power values are within [210, 1751] kW, making "0" a unique value to be distinguishable.

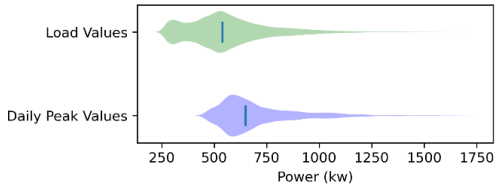

Fig. 3. Distribution of the load values and daily peak values.

To address the influence of temperature on load, we include the normalized ambient temperature profile data, designated as $T$, as an additional modality input to assist in the recovery of missing load data. $X^m$ and $T$ are subjected to normalization based on the highest and lowest annual power value and temperatures respectively, ensuring that the normalized power and temperature values range from 0 to 1. Then, $X^m$ and $T$ are rescaled to the [0, 200] range.

### 2) Encoding Smart Meter and Temperature data to Words

ChatGPT is primarily a large language model, so our approach involves converting load and temperature values into words that can be easily interpreted by ChatGPT. We propose an encoding method to transform smart meter data and temperature values into words.

After data preprocessing, both load and temperature data are represented by integers between 0 and 200. We use a five-digit ternary code to represent each load or temperature value, with each digit having three possible values: "0", "1", and "2". This five-digit ternary code can represent $3^5 = 243$ different numbers, which is sufficient for our range. We then convert the ternary code into a "word" by replacing "0", "1", and "2" with "L", "M", and "H", respectively. Missing values are encoded as "OOOOO" to distinguish them from actual data.

To further explore time alignment capabilities, we combine the load and temperature encodings at each timestamp into a single long "word", where the first five letters represent the load value, and the last five letters represent the temperature value.

An illustration of this load-temperature to "word" encoding is shown in Figure 4. The example includes a sequence of three time points with a missing value in the middle. The load and temperature sequences are encoded into a long "word" sequence. This sequence is then fed into ChatGPT using a proper prompting strategy (introduced in the next section) to obtain a restored load "word" sequence. Finally, the restored load value is generated by decoding the restored "word", which is the inverse process of encoding.

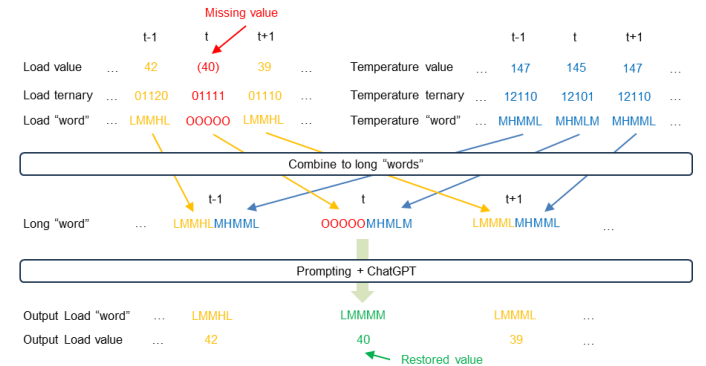

Fig. 4. An illustration of load-temperature to "word" encoding method.

In the simulation section, we will discuss the performance across different scenarios: "encoded" versus "not-encoded",

and "load-temperature-as-a-single-word" versus "load-temperature-as-separate-words".

*3) Prompt Engineering*

Prompt engineering is a methodology employed in the field of natural language processing (NLP) to design effective prompts for language models like GPT-3.5. It involves crafting specific instructions or queries that guide the model to generate desired outputs or responses. This process requires understanding the capabilities and limitations of the model, as well as the nuances of the task or domain for which prompts are being created. Effective prompt engineering can significantly improve the performance and applicability of language models across various tasks, including missing data restoration, demand response baseline estimation, and more.

In the power system demand response baseline estimation task, as shown in Figure 5(a), we structure the prompt in a multi-turn chat format. The conversation follows the format defined by OpenAI [45], consisting of a list of messages where each message has a role and content. The "user" is the person interacting with ChatGPT, and the "assistant" represents ChatGPT. Initially, the user informs the assistant about the task and explains how the data is defined. Then, the user provides the encoded data as requested by the assistant. Finally, the assistant estimates the missing data and outputs the complete load encodings, which can be decoded to generate the complete load profile.

```
{"messages":
    [{"role": "user",
      "content":"Given a load profile with missing segments and a complete"
                "daily temperature profile, estimate the missing portions of"
                "the load profile. The load and temperature data are provided"
                "jointly. Each value for load and temperature is encoded as a"
                "ten-digit word in ternary format. The first and last five"
                "digits represent load and temperature, respectively. Missing"
                "load values are represented by 00000. Please provide the"
                "estimated load profile in the same format and length."},
     {"role": "assistant",
      "content": "What's the encoded data?"},
     {"role": "user",
      "content": long_encodings},
     {"role": "assistant",
      "content": completion}
    ]
}
```
(a)

```
{"messages":
    [{"role": "user",
      "content":"Given a load profile with missing segments and a complete"
                "daily temperature profile, estimate the missing portions of"
                "the load profile. The load and temperature data are provided"
                "separately and exhibit a correlation. Each value for load or"
                "temperature is encoded as a five-digit word in ternary format."
                "Missing load values are represented by 00000. Please provide"
                "the estimated load profile in the same format and length."},
     {"role": "assistant",
      "content": "What's the encoded load profile?"},
     {"role": "user",
      "content": load_encodings},
     {"role": "assistant",
      "content": "What's the encoded temperature?"},
     {"role": "user",
      "content": temperature_encodings},
     {"role": "assistant",
      "content": completion}
    ]
}
```
(b)

Fig. 5. Prompting methods used in missing data restoration. (a) load-temperature-as-a-single-word. (b) load-temperature-as-separate-words.

Combining load and temperature into a single long encoding is optional. We can also provide load and temperature encodings separately, as illustrated in Figure 5(b). The comparison between these two prompting methods is discussed in the simulation results section.

A multi-turn chat, as shown in Figure 5, serves as a data sample for fine-tuning. A collection of such data samples constitutes a fine-tuning dataset, which is used to fine-tune ChatGPT. Once fine-tuning is complete, ChatGPT will be able to understand the inherent correlations between the missing parts and the rest of the load data, enabling it to perform missing data restoration and demand response baseline estimation. For testing purposes, the testing set is prepared in the same format, with the last message in each multi-turn chat removed.

*4) OpenAI Fine Tuning*

Fine-tuning is a powerful technique because it leverages the knowledge and general language understanding already present in the pre-trained model, allowing for efficient adaptation to new tasks or domains with relatively small amounts of additional data. It has been widely used to customize language models for a wide range of applications.

Fine-tuning in the context of OpenAI's language models involves adjusting a pre-trained model on a specific dataset or task to improve its performance for that particular use case. When fine-tuned, the model's parameters are updated based on new data, allowing it to adapt to the specifics of the task or domain.

For example, with GPT-3.5, fine-tuning involves providing additional training data related to a specific task, such as missing data restoration or demand response baseline estimation, and then adjusting the model's parameters through further training on this data. This process helps the model learn task-specific patterns and nuances, ultimately enhancing its ability to perform well on the given task.

To create a fine-tuning job with OpenAI, we first upload the fine-tuning dataset created in the previous section, then initiate the fine-tuning job through the Fine-tuning User Interface [46] or programmatically. The OpenAI platform will then fine-tune the specified model using the provided dataset and generate the fine-tuned model. In our fine-tuning job, we experimented with different numbers of samples in the dataset. These data sizes are relatively small compared to those used for training a traditional machine learning model from scratch. The performance evaluation of different fine-tuning data sizes will be presented in the simulation results section.

*D. 2nd Stage Fine-tuning*

In the second stage, we use an even smaller amount of data from the target user, whose pattern is similar to the 10 users used in the first stage, to further fine-tune the model obtained from the first stage.

Fine-tuning is a powerful technique because it leverages the knowledge and general language understanding already present in the pre-trained model, enabling efficient adaptation to new tasks or domains with relatively small amounts of additional data. The first stage fine-tunes ChatGPT from a general language model to a specific missing data restoration model for a group of users. The second stage then adapts this model for the target user, effectively narrowing the gap between a general model and a highly specialized model, making the fine-tuning process smoother. If we were to fine-tune ChatGPT directly with data from the target user, the limited data availability would prevent the model from performing well.

The data preprocessing, encoding, and prompt engineering techniques used in this stage are the same as those in the first stage. We also experimented with different numbers of samples for the second stage fine-tuning, resulting in varying levels of missing data restoration accuracy. Additionally, we attempted to fine-tune ChatGPT directly with data from the target user. The results of these experiments will be presented in the simulation results section.

*E. Evaluation Metrics*

The performance metrics used for evaluating the accuracy of the restored data segments are calculated as

$$MPE = \frac{1}{K}\sum_{t=1}^{K}\frac{|\hat{x}_t^m - x_t^m|}{x_t^{event}} \quad (5)$$

$$RMSE = \sqrt{\frac{1}{K}\sum_{t=1}^{K}(\hat{x}_t^m - x_t^m)^2} \quad (6)$$

$$EGYE = \frac{|\sum_{t=1}^{K}\hat{x}_t^m - \sum_{t=1}^{K}x_t^m|}{\sum_{t=1}^{K}x_t^{event}} \quad (7)$$

where $K$ is the total number of data points in the MDS, and $\hat{x}$ represents the restored data segment. These indices offer insights into different aspects of errors, including point-to-point errors and discrepancies in total energy consumption.

## III. SIMULATION RESULTS

In this section, we assess the ability of the fine-tuned large language model to recover missing data segments (MDS) under different scenarios and compare it with the benchmark method BERT-PIN [28]. The evaluation encompasses diverse fine-tuning scenarios and variations in training data volumes, providing valuable insights into fine-tuning Large Language Models for tasks in power system data analytics. Lastly, we discuss applying the proposed method to other power system data analysis tasks, such as load forecasting.

*A. Data Preparation*

The load profiles used in this study consists of 15-minute resolution smart meter data obtained from 11 users in North Carolina in summer 2018. The corresponding temperature data is downloaded from the National Oceanic and Atmospheric Administration (NOAA) [47] website and is used as a second modality input.

These load profiles are aligned with the temperature profile and then normalized based on the overall load and temperature peaks, respectively. Then, we partition the profiles into daily (96 data points) profiles. Each missing load data segment is consistently set at 4 hours (16 data points), a choice guided by the observation that around 70% of missing load data segments in actual utility data are less than 4 hours in duration. It's important to note that there are no missing data segments in the temperature profile. The dataset is divided into an 80% training set and a 20% testing set.

*B. Evaluation of 1st Stage Fine-tuning*

In the first stage of fine-tuning, the standard GPT-3.5 model is fine-tuned using data collected from 10 users. These users are evenly distributed between the fine-tuning and testing datasets. Subsequently, the fine-tuned GPT model is utilized to restore a 4-hour missing window in a daily load profile. BERT-PIN [28] is chosen as the benchmark method due to its superior performance in power system missing data restoration compared to other existing methods. Figure 6 presents several restored daily load profiles. Upon visual inspection, it is evident that the fine-tuned GPT model (represented by the red lines) achieves comparable performance with BERT-PIN (represented by the blue lines). The missing data segments restored by both methods closely resemble the ground truth.

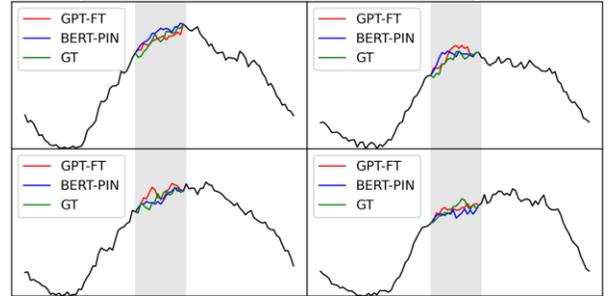

Fig. 6. Examples of missing data restoration by fine-tuned model and benchmark model.

To quantitatively evaluate the accuracy of the fine-tuned model, we explored various scenarios. Before delving into these scenarios, it's important to clarify some concepts. A sample refers to a multi-turn chat message depicted in Figure 6, representing one daily load profile. "Advanced prompt" is a prompt method introduced in section II.C.3, while "non-advanced prompt" implies the absence of detailed explanation information in the message, making it challenging for GPT to comprehend the encoding strategy and data features. "Discard encoding" refers to replacing load and temperature encodings with integers. "Abnormal days" are those when there is a sudden drop in temperature, which could affect the quality of the fine-tuning data, particularly during summer.

The experiment comprises seven scenarios:

**Scenario 1:** Fine-tune with 128 samples, using a non-advanced prompt, and combining load data and temperature data in the input. The encoding method is employed, and abnormal days are not removed.

**Scenario 2:** Similar to scenario 1, but with the number of samples increased to 256.

**Scenario 3:** Similar to scenario 2, but with the number of samples further increased to 512.

**Scenario 4:** Similar to scenario 3, but adopting the advanced prompt method introduced in section II.C.3.

**Scenario 5:** Similar to scenario 4, but with load encoding and temperature encoding separated in the input, as shown in Figure 5(b).

**Scenario 6:** Similar to scenario 5, but removing the encoding part, and directly using integers representing load and temperature values as input.

**Scenario 7:** Similar to scenario 6, but removing abnormal days from the fine-tuning dataset.

TABLE II
MISSING DATA RESTORATION IN DIFFERENT SCENARIOS

| | # of samples | Advanced prompt | Separate Load & temperature | Discard Encoding | Remove abnormal days | Errors (%) | | |
|---|---|---|---|---|---|---|---|---|
| | | | | | | MPE | RMSE | EGYE |
| Scenario 1 | 128 | N | N | N | N | 5.609 | 4.513 | 4.303 |
| Scenario 2 | 256 | N | N | N | N | 4.609 | 3.816 | 3.652 |
| Scenario 3 | 512 | N | N | N | N | 4.59 | 3.795 | 3.547 |
| Scenario 4 | 512 | Y | N | N | N | 3.806 | 3.027 | 2.987 |
| Scenario 5 | 512 | Y | Y | N | N | 3.266 | 2.656 | 2.469 |
| Scenario 6 | 512 | Y | Y | Y | N | 2.48 | 2.029 | 1.639 |
| Scenario 7 | 512 | Y | Y | Y | Y | **2.221** | **1.977** | **1.443** |
| BERT-PIN | ~220K | - | - | - | - | 1.612 | 0.699 | 0.887 |

Based on the results presented in Table III, the following observations were made:

- Across scenarios 1-3, using more samples in the first stage fine-tuning leads to higher accuracy. Increased data coverage enables the fine-tuned model to better capture various load patterns, thus enhancing its capability in missing data restoration. The selection of 512 samples was based on data availability, with each user contributing about 50 samples from the 86 summer days, leaving the remainder for testing purposes. Thus, 512 is selected as the number of samples in the following scenarios.
- Scenario 4, which employs an advanced prompt, yields lower errors. The detailed information about the task description, encoding strategy, and output constraints provided in the advanced prompt aids GPT in understanding the correlations between load and temperature as well as contextual information within load sequences, thereby improving the model's performance.
- Scenario 5, where the load encoding and temperature encoding are presented in a two-turn chat, instead of in one single long encoding, simplifies the encoding format. Because GPT model does not need to identify which part of the long encoding represents load and which part represents temperature. Therefore, scenario 5 further decreases the restoration error.
- Inspired by the success of scenario 5, scenario 6 removes the encoding mechanism entirely, replacing load and temperature encodings with normalized integers. This results in further performance improvement, indicating that ChatGPT can comprehend direct numbers better than encoded "words." Because these "words" have not appeared in ChatGPT's training dataset, it's hard to make ChatGPT understand it by fine-tuning with a small amount of data.
- In scenario 7, abnormal days with sudden temperature drops are removed from the fine-tuning dataset. This mitigates the adverse effect of temperature drops on power consumption, which makes the model's parameters be updated to another direction. Consequently, the removal of abnormal days leads to a further increase in missing data restoration accuracy.
- Despite the decreasing errors observed from scenario 1 to scenario 7, the performance remains slightly below that of the benchmark model BERT-PIN [28]. BERT-PIN is a BERT-based model that has been trained from scratch using a significantly larger dataset, comprising approximately 220,000 samples collected from 8,000 users over three years, with a training duration of 6-8 hours. In contrast, the proposed fine-tuned model only requires 512 samples collected from 10 users over three months, with a fine-tuning time less than 1 hour. Despite using fewer resources, the fine-tuned model achieves acceptable performance. Moreover, further performance improvements are expected in the second stage of fine-tuning.

C. *Evaluation of 2nd Stage Fine-tuning*

In the first stage, GPT-3.5 undergoes fine-tuning with data collected from 10 users (user0 – user9), resulting in the generation of the fine-tuned model GPT-FT-1. Subsequently, in the second stage, GPT-FT-1 undergoes further fine-tuning with data collected from a target user, distinct from the users utilized in the first stage (user10 - user19). It's important to note that only one target user is employed in the second stage fine-tuning, with the experiment being repeated using 10 different target users to assess the reliability of the fine-tuning strategy.

In the second stage fine-tuning, varying numbers of samples ranging from 10 to 50 are utilized to investigate the requisite sample size. This exploration is particularly advantageous in scenarios where suitable training data is scarce. Additionally, the performance of the GPT-FT-1 model without the second stage fine-tuning is evaluated on the target user to demonstrate the transferability of the first stage fine-tuned model. Furthermore, attempts are made to fine-tune the original GPT-3.5 model solely with data from the target user. In this scenario, 68 samples from the target user are utilized for fine-tuning, with the remaining 18 samples reserved for testing purposes. The setup of this scenario, depicted in Figure 7, serves to illustrate the necessity of the proposed two-stage fine-tuning strategy, particularly in cases where suitable training data is scarce.

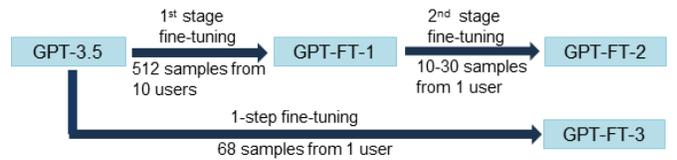

Fig. 7. An illustration of fine-tuning the original GPT-3.5 model solely with data from the target user.

Based on the results presented in Table III, the following observations were made:

- Contrary to common intuition, the performance of the second stage fine-tuned model does not improve with an increase in training data. Instead, optimal accuracy is achieved with 10 to 30 fine-tuning samples, while adding more samples (40 or 50) leads to decreased performance. This suggests that the GPT-FT-1 model, following the first stage fine-tuning, has grasped the underlying principles of the missing data restoration task in general user cases. Consequently, only a small number of samples are required for effective transfer of knowledge from general users to a specific target user, with additional data being redundant and detrimental to accuracy.

TABLE III
ERRORS OF SECOND STAGE FINE-TUNING

| Target Users | Errors (%) | GPT-FT-2 fine-tuned with different numbers of samples | | | | | GPT-FT-1 (W/O 2nd stage) | | GPT-FT-3 (W/O 1st stage) |
|---|---|---|---|---|---|---|---|---|---|
| | | 10 | 20 | 30 | 40 | 50 | Test on target user | Test on user0-9 | |
| User10 | MPE | 2.245 | 2.277 | 2.428 | 2.395 | 2.361 | 2.497 | 2.221 | 2.983 |
| | RMSE | 1.988 | 1.969 | 2.014 | 2.094 | 2.140 | 2.182 | 1.977 | 2.772 |
| | EGYE | 1.501 | 1.513 | 1.599 | 1.604 | 1.637 | 1.843 | 1.443 | 2.461 |
| User11 | MPE | 2.206 | 2.168 | 2.416 | 2.352 | 2.549 | 2.77 | 2.221 | 3.216 |
| | RMSE | 1.911 | 1.887 | 1.957 | 2.008 | 2.164 | 2.438 | 1.977 | 2.837 |
| | EGYE | 1.499 | 1.453 | 1.534 | 1.649 | 1.692 | 2.103 | 1.443 | 2.524 |
| User12 | MPE | 2.276 | 2.453 | 2.483 | 2.533 | 2.676 | 2.441 | 2.221 | 3.068 |
| | RMSE | 1.962 | 1.997 | 2.016 | 2.147 | 2.163 | 2.206 | 1.977 | 2.649 |
| | EGYE | 1.524 | 1.517 | 1.567 | 1.599 | 1.643 | 1.784 | 1.443 | 2.581 |
| User13 | MPE | 2.382 | 2.448 | 2.565 | 2.604 | 2.412 | 2.503 | 2.221 | 2.979 |
| | RMSE | 2.009 | 2.018 | 2.097 | 2.105 | 2.166 | 2.263 | 1.977 | 2.729 |
| | EGYE | 1.587 | 1.589 | 1.535 | 1.673 | 1.749 | 1.958 | 1.443 | 2.441 |
| User14 | MPE | 2.318 | 2.248 | 2.558 | 2.48 | 2.336 | 2.673 | 2.221 | 3.110 |
| | RMSE | 1.997 | 1.948 | 2.021 | 2.134 | 2.196 | 2.273 | 1.977 | 2.811 |
| | EGYE | 1.472 | 1.499 | 1.514 | 1.731 | 1.887 | 2.094 | 1.443 | 2.687 |
| User15 | MPE | 2.314 | 2.273 | 1.999 | 2.088 | 2.116 | 2.528 | 2.221 | 2.914 |
| | RMSE | 1.964 | 1.897 | 1.846 | 1.943 | 2.155 | 2.364 | 1.977 | 2.761 |
| | EGYE | 1.507 | 1.414 | 1.634 | 1.697 | 1.805 | 1.833 | 1.443 | 2.416 |
| User16 | MPE | 2.186 | 2.302 | 2.418 | 2.336 | 2.237 | 2.767 | 2.221 | 3.131 |
| | RMSE | 1.885 | 1.934 | 2.149 | 2.261 | 2.381 | 2.419 | 1.977 | 2.828 |
| | EGYE | 1.455 | 1.393 | 1.620 | 1.678 | 1.803 | 1.972 | 1.443 | 2.338 |
| User17 | MPE | 2.453 | 2.323 | 2.681 | 2.698 | 2.536 | 2.625 | 2.221 | 3.401 |
| | RMSE | 2.012 | 2.244 | 2.198 | 2.265 | 2.337 | 2.397 | 1.977 | 2.945 |
| | EGYE | 1.572 | 1.461 | 1.628 | 1.724 | 1.796 | 2.007 | 1.443 | 2.367 |
| User18 | MPE | 2.466 | 2.449 | 3.019 | 2.466 | 2.777 | 2.92 | 2.221 | 3.274 |
| | RMSE | 2.194 | 2.130 | 2.248 | 2.555 | 2.730 | 2.615 | 1.977 | 2.716 |
| | EGYE | 1.419 | 1.527 | 1.699 | 1.647 | 1.933 | 2.118 | 1.443 | 2.592 |
| User19 | MPE | 2.693 | 2.598 | 2.474 | 2.662 | 2.856 | 2.644 | 2.221 | 3.392 |
| | RMSE | 2.248 | 2.267 | 2.101 | 2.377 | 2.409 | 2.529 | 1.977 | 2.779 |
| | EGYE | 1.623 | 1.584 | 1.770 | 1.764 | 2.008 | 2.177 | 1.443 | 2.486 |

- Testing GPT-FT-1 on the target user yields increased errors compared to testing it on users0-9. This discrepancy arises because GPT-FT-1 is fine-tuned with data from users0-9, resulting in better performance for these users. However, even when tested on data from a new user, GPT-FT-1 still achieves acceptable accuracy, indicating that the first stage fine-tuning endows GPT-FT-1 with a general understanding of the missing data restoration problem, thereby possessing potential for further fine-tuning to enhance performance.
- Compared to testing GPT-FT-1 without the second stage fine-tuning on target user (second column from the right of Table III), the second stage fine-tuned model GPT-FT-2 exhibits improved accuracy for all target users. This underscores the effectiveness of the second stage fine-tuning in tailoring the model to the target data, successfully transferring the general understanding of missing data restoration to the specific user context with minimal labeled examples.
- Direct fine-tuning of the original GPT-3.5 model with 68 samples from the target user results in inferior performance compared to all other cases. While increased training data would likely improve accuracy significantly, such an approach is impractical in scenarios of data scarcity. In contrast, the two-stage fine-tuning strategy addresses this challenge by leveraging data from other users over a shorter timeframe, circumventing the need for extensive data collection from the target user over a prolonged period.

### D. Cost Analysis

The fine-tuning strategy proposed in this paper is executed through the OpenAI API. We upload the prepared fine-tuning data onto the OpenAI platform and initiate a fine-tuning task. The platform then proceeds to fine-tune the specified model with the uploaded data, generating the fine-tuned model. The cost associated with this process varies depending on the number of samples utilized in the fine-tuning process.

As depicted in Figure 8(a), employing more samples results in a higher number of tokens trained by the platform. According to pricing information sourced from OpenAI [48], the cost amounts to $8 per 1M tokens trained. Consequently, the training cost for different sample sizes is illustrated in Figure 8(b). For instance, fine-tuning with a dataset comprising 512 samples would incur approximately $13 in training costs. It's important to note that this cost solely pertains to the fine-tuning process; testing costs are calculated separately but are relatively lower compared to training expenses, hence not depicted in

Figure 8.

One of the key advantages of utilizing fine-tuning is its time efficiency. Fine-tuning 512 samples takes merely 1 hour to achieve acceptable results. This stands in stark contrast to the hours required to train a specific model from scratch. Therefore, fine-tuning represents a trade-off between marginal performance improvements and computational resource consumption.

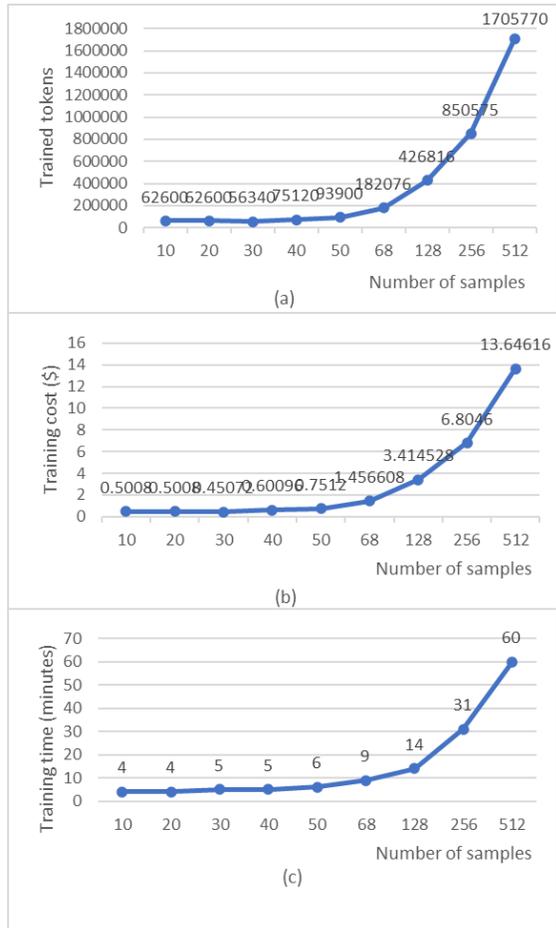

Fig. 8. The cost of fine-tuning versus number of samples.

## IV. Conclusion

In conclusion, this paper introduces a novel approach using fine-tuned LLMs for load profile analysis, exemplified by the missing data restoration task. We adapted the pre-trained GPT-3.5 model for power system load analytics through a two-stage fine-tuning process. Our empirical results demonstrate that the fine-tuned model restores missing data with accuracy comparable to advanced models like BERT-PIN.

Key insights include the effectiveness of using a minimal number of fine-tuning samples, which underscores the efficiency of few-shot learning. Additionally, advanced prompt engineering and separate encoding of load and temperature data significantly enhance model performance. The fine-tuning strategy proved to be cost-effective and time-efficient, presenting a viable alternative to training models from scratch, especially in data-limited and resource-scarce settings.

Future work from this research will focus on extending the application of LLMs to additional power system load analysis tasks, including load forecasting, load disaggregation, customer segmentation, and synthetic data generation.